\documentclass[conference]{IEEEtran}
\usepackage[plain]{algorithm}
\usepackage{algpseudocode}
\algdef{SE}[DOWHILE]{Do}{doWhile}{\algorithmicdo}[1]{\algorithmicwhile\ #1}%
\algnewcommand{\var}{\textit}

\usepackage{tikz}
\usetikzlibrary{plotmarks}
\usepackage{pgfplots}
\pgfplotsset{compat=newest}

\newlength\figureheight 
\newlength\figurewidth 

\usepackage{subfig}
\usepackage{enumitem}
\usepackage{multirow}
\usepackage{graphicx}
\usepackage{amsmath}

\usepackage{stfloats}
\usepackage{bm}	

\newcommand\DV{D\hspace{-0.1em}V}
\newcommand\TP{T\hspace{-0.1em}P}
\newcommand\TN{T\hspace{-0.1em}N}
\newcommand\agg{\text{agg}}

\makeatletter
\renewcommand{\ALG@name}{Alg.}
\makeatother
\usepackage{caption}
\DeclareCaptionLabelFormat{alglabel}{\csname ALG@name\endcsname#1 #2}
\newsavebox{\ieeealgbox}
\newenvironment{boxedalgorithmic}
{\begin{lrbox}{\ieeealgbox}
		\begin{minipage}{\dimexpr\columnwidth-2\fboxsep-2\fboxrule}
			\begin{algorithmic}}
			{\end{algorithmic}
		\end{minipage}
	\end{lrbox}\noindent\fbox{\usebox{\ieeealgbox}}}

\makeatletter
\newcommand\fs@norules{\def\@fs@cfont{\bfseries}\let\@fs@capt\floatc@ruled
	\def\@fs@pre{}%
	\def\@fs@post{}%
	\def\@fs@mid{\kern3pt}%
	\let\@fs@iftopcapt\iftrue}
\makeatother
\floatstyle{norules}
\restylefloat{algorithm}

\usepackage{booktabs}
\usepackage{makecell}

\renewcommand{\baselinestretch}{0.981}
\begin{document}
\title{Active Learning for One-Class Classification Using Two One-Class Classifiers}

\author{
	\IEEEauthorblockN{Patrick Schlachter and Bin Yang}
	\IEEEauthorblockA{Institute of Signal Processing and System Theory\\University of Stuttgart, Germany\\\{patrick.schlachter, bin.yang\}@iss.uni-stuttgart.de}}

\maketitle

\begin{abstract}
This paper introduces a novel, generic active learning method for one-class classification.
Active learning methods play an important role to reduce the efforts of manual labeling in the field of machine learning. Although many active learning approaches have been proposed during the last years, most of them are restricted on binary or multi-class problems.
One-class classifiers use samples from only one class, the so-called target class, during training and hence require special active learning strategies. The few strategies proposed for one-class classification either suffer from their limitation on specific one-class classifiers or their performance depends on particular assumptions about datasets like imbalance.
Our proposed method bases on using two one-class classifiers, one for the desired target class and one for the so-called outlier class. It allows to invent new query strategies, to use binary query strategies and to define simple stopping criteria. Based on the new method, two query strategies are proposed.
The provided experiments compare the proposed approach with known strategies on various datasets and show improved results in almost all situations.
\end{abstract}

\IEEEpeerreviewmaketitle

\section{Introduction}
A general challenge of supervised machine learning is the acquisition of hand-labeled samples.
Since labeling often requires human experts, it is a time-consuming and costly task. Therefore, it is highly desirable to reduce the manual labeling effort. This is also the case for one-class classification.

One-class classification tries to recognize samples of one given class, the target class, against samples belonging to the other class, the outlier class. Its specialty lies in the fact that only samples from the target class are used to train a one-class classifier (OCC).
Samples of the outlier class are rarely available or, if they are available, not representative to describe different types of outliers and thus cannot be used.
In contrast to binary classification, an OCC does not try to separate samples from two classes but encompasses the training samples of the target class.
During the evaluation of an OCC, samples from both classes are necessary of course.
An overview of one-class classification and various OCCs was shown in \cite{KhanM13}, \cite{Tax01}.

Although one-class classification was first motivated by the absence of outlier samples during training, it is also deployed in other scenarios. For example, OCCs are used in some binary classification applications whose training data is either imbalanced or only representative of one of both classes. An example of the latter case is fault detection where possible types of faults are unknown in advance.

Despite requiring only samples from one class during training, the problem of a high labeling effort is still present. An expert, a so-called \emph{oracle}, has to manually select target samples from a pool of unlabeled data which are subsequently used to train an OCC. Regarding one-class classification, the problem is aggravated by the fact that many one-class applications have to strictly avoid false negative (FN) errors, i.e. outliers classified as targets. Consequently, decision boundaries of OCCs are normally chosen tight to training samples of the target class. This can lead to many false positive (FP) errors, i.e. targets classified as outliers, at which human experts typically have to take a closer look. If a target is found, it can be used to iteratively retrain the OCC.

Active learning is a technique to tackle the problem of high labeling efforts by querying only the samples that are most valuable for the training of a classifier. There exist various query strategies \cite{Settles09} which differ in the interpretation of the term "valuable".
However, most of the proposed query strategies are limited to binary or multi-class classifiers. Active learning for one-class classification is more challenging because the decision boundary of an OCC is only supported by samples from one class. Hence, labeled outliers, i.e. samples that the oracle marked as outliers, cannot be used for iterative training.

Some former works proposed active learning strategies specially tailored to one-class classification.
In general, they are a trade-off between \emph{exploration} and \emph{exploitation} of the feature space. Querying a sample in a non-sampled area of the feature space corresponds to exploration and helps to extend the decision boundary to the non-sampled area. In contrast, selecting a sample in a sampled area of the feature space contributes to exploitation and helps to refine the decision boundary.
The query strategy uncertainty sampling presents those samples to an oracle about which a classifier is the most uncertain \cite{Lewis94}. Juszczak et al. adopted uncertainty sampling to one-class classification by introducing confidence values \cite{Juszczak03}. They proposed four query strategies from which the "lh" strategy works best for almost all experiments. It presents estimated targets with low confidence and estimated outliers with high confidence to the oracle.
In contrast to uncertainty sampling, Barnab\'e-Lortie et al. proposed to query samples about which an OCC is the most certain that they are outliers \cite{Barnabe-Lortie15}.
Ghasemi et al. presented a different approach which makes use of the distribution of target and unlabeled samples and does not consider the classification results of an OCC for active learning. Based on kernel density estimation, they proposed the two query strategies expected margin sampling and entropy sampling \cite{Ghasemi112}.
There exist further query strategies which are not considered in this paper due to their limitation to special OCCs or high computational cost. Some of them are based on kernel density estimation \cite{Ghasemi11, Juszczak06}, support vector data description \cite{Görnitz2009, Liu08} or the variation in label assignments \cite{Juszczak06}.
Finally, it was shown in \cite{Ghazel17} that active learning for one-class classification can be improved by applying dimensionality reduction.

Most of the presented active learning strategies for OCCs have one major drawback: If a sample is labeled as an outlier by the oracle, both the OCC and the active learning strategy receive no information gain. During training, the reason for having no information gain obviously lies in the nature of one-class classification. However, labeled outliers can improve active learning.

\begin{figure}[!b]
	\centering
	\begin{minipage}{0.48\textwidth}
		\begin{algorithm}[H]
			\caption{Proposed active learning method for one-class classification based on two OCCs}
			\label{alg:proposed_al_method}
			\normalsize
			\begin{boxedalgorithmic}[1]
				\State given: target set $\mathcal{T}$, outlier set $\mathcal{O}$, unlabeled set $\mathcal{U}$
				\State $\bm{x}_i=\emptyset$, $y_i = \text{target}$, $\DV^t_u=\DV^o_u=1$ $\forall$ $u=1, ..., |\mathcal{U}|$
				
				\Loop
				\If {$y_i$ = target}
				\State add sample $\bm{x}_i$ to $\mathcal{T}$
				\State retrain target OCC with $\mathcal{T}$
				\State test target OCC on $\mathcal{U}$ $\Rightarrow$ $\{ \DV^t_u, u=1, ..., |\mathcal{U}|\}$
				\Else
				\State add sample $\bm{x}_i$ to $\mathcal{O}$
				\State retrain outlier OCC with $\mathcal{O}$
				\State test outlier OCC on $\mathcal{U}$ $\Rightarrow$ $\{ \DV^o_u, u=1, ..., |\mathcal{U}|\}$
				\EndIf
				\State $\bm{x}_i= \arg\max\limits_u \{ \agg(\DV^t_u, \DV^o_u), u=1, ..., |\mathcal{U}|\}$
				\State query label $y_i$ of sample $\bm{x}_i$ from oracle
				\State remove sample $\bm{x}_i$ from $\mathcal{U}$
				\EndLoop
			\end{boxedalgorithmic}	
		\end{algorithm}
	\end{minipage}
\end{figure}

In this paper, we propose a new active learning method for one-class classification which is able to take labeled outliers into account. Our method is generic in the sense that it enables to define various query strategies. Furthermore, it is universally applicable, i.e. it is not limited to specific OCCs, and it is adjustable in multiple manners like the trade-off between exploration and exploitation. Finally, the proposed method does not rely on specific assumptions on a dataset or on the distribution of the outlier class. Based on the generic method, two query strategies are proposed.

During our experiments, we made use of a popular one-class classifier: the one-class support vector machine (OCSVM) proposed by Sch\"olkopf et al. \cite{Scholkopf2001}. It is based on the maximization of the margin between a hyperplane and the origin of the feature space. The solution of an estimated class label $\hat{y}$ for a given feature vector $\bm{x}$ and a training set $\{\bm{x}_i\}_{i=1}^n$ is given by	
\begin{equation}
\hat{y}(\bm{x})=\text{sgn}\Bigl(\underbrace{\sum_{i=1}^n\alpha_i K(\bm{x},\bm{x}_i)-\rho}_{\DV}\Bigr).
\end{equation}
The value of the discriminant function whose sign defines the class affiliation is called \emph{decision value} $\DV$. It is based on Lagrange multipliers $\alpha_i$, an offset $\rho$ and a kernel $K$. 
As our method is not limited to OCSVMs, decision values may emerge from other OCCs. In general, decision values are negative for samples that belong to the training class and positive for the other class. The higher the absolute value of a decision value is, the more certain an OCC is about the estimated class affiliation.

The remaining parts of this paper are structured as follows: In the next section, we present our main contribution, a novel active learning method for one-class classification. Section 3 evaluates the method on multiple datasets using an OCSVM as OCC and compares it to other query strategies. Finally, we will conclude our work in Section 4.

\section{Proposed method}
The proposed generic active learning method for one-class classification is based on two OCCs: one for the target class and one for the outlier class. The target-class classifier is used for classification and both OCCs are used for active learning.

\subsection{Basic idea}	
The basic idea behind this method is to make use of labeled outliers during active learning. As mentioned in the introduction, OCCs are also deployed in applications where outliers occur frequently but are not necessarily representative of the whole distribution of the outlier class. Therefore, switching to a binary classifier after obtaining some labeled outliers is not always reasonable. Indeed, binary classification carries the risk of producing many FN errors, whereas an OCC leads to mostly FP errors which is more desirable in many one-class applications.
Although outlier samples are insufficient for classification, an OCC trained on them can support active learning.

We notice that combining OCCs is well known for both multi-class classification \cite{Hadjadji14, Krawczyk15} and to improve one-class classification in terms of performance and robustness \cite{Tax012}. However, for the purpose of active learning, joining two one-class classifiers is novel.

\subsection{Algorithm}
Alg. \ref{alg:proposed_al_method} outlines the proposed active learning process for an OCC.
$y_i$ denotes the class label of the previously added sample $\bm{x}_i$ and is initialized to the target label. If $y_i$ is a target, then the target OCC is retrained on the increased target set $\mathcal{T}$ and is used to test the samples in the unlabeled set $\mathcal{U}$. If $y_i$ is an outlier, then the outlier OCC is retrained on the increased outlier set $\mathcal{O}$ and is used again to test all unlabeled samples. Together with the respective second stored decision boundary, this results in two decision values $\DV_u^t$ and $\DV_u^o$ for each unlabeled sample which are subsequently aggregated using an \emph{aggregation function} $\agg$.
Finally, the sample with maximal aggregate is presented to an oracle.
Corresponding to other active learning approaches, this querying sequence is repeated until a defined stopping criterion is fulfilled.

The method has several advantages. First, it is able to consider labeled outliers during active learning. In addition, only one of both OCCs has to be retrained per query step which makes it computationally efficient. Moreover, the generic method can be adapted by choosing different aggregation functions $\agg$. Further adjustments are conceivable by using different hyperparameters for both OCCs.
	
The following two subsections propose two meaningful aggregation functions which result in the query strategies \emph{exploration sampling} and \emph{similarity sampling}.

\begin{figure*}[!t]
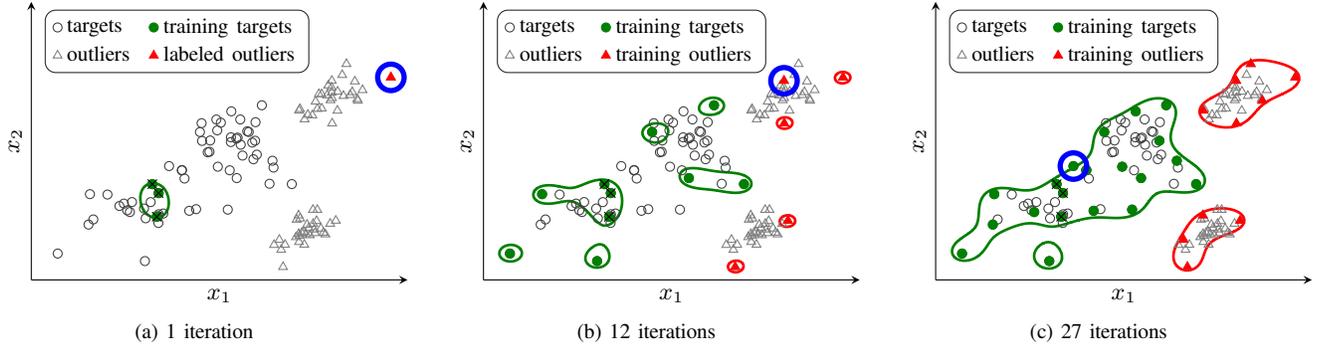

	\centering
	\subfloat[1 iteration]{
		\setlength\figureheight{3.7cm} 
		\setlength\figurewidth{0.29\textwidth}
		\input{figures/clusters_exploration_sampling_1_multiple.tikz}
		\label{fig:exploration_sampling_clusters_1_iteration}
	}
	\hfil
	\subfloat[12 iterations]{
		\setlength\figureheight{3.7cm} 
		\setlength\figurewidth{0.29\textwidth}
		\input{figures/clusters_exploration_sampling_12_multiple.tikz}
		\label{fig:exploration_sampling_clusters_12_iterations}
	}
	\hfil
	\subfloat[27 iterations]{
		\setlength\figureheight{3.7cm} 
		\setlength\figurewidth{0.29\textwidth}
		\input{figures/clusters_exploration_sampling_27_multiple.tikz}
		\label{fig:exploration_sampling_clusters_27_iterations}
	}
	\caption{Exploration sampling on artificial dataset after 1, 12 and 27 iterations. The three initial training targets are marked by crosses. Decision boundaries of target OCC and outlier OCC are depicted as green and red lines, respectively. The currently queried sample is marked by a thick blue circle.}
	\label{fig:exploration_sampling_clusters}
\end{figure*}

\subsection{Exploration sampling}\label{subsec:exploration_sampling}
The query strategy \emph{exploration sampling} is based on the generic method from above and uses the following aggregation function:
\begin{equation}
	\agg_{\text{exp}} = \DV^t \cdot \DV^o.
\end{equation}
In contrast to outlier sampling \cite{Barnabe-Lortie15}, exploration sampling presents samples to an oracle that lie not only furthermost from the decision boundary of the target OCC but also furthermost from the decision boundary of the outlier OCC.
This leads to a high exploration which is crucial in many one-class applications that either have to be operational based on only a few training samples or in which asking an oracle is expensive.
The risk of presenting many true outliers to an oracle is reduced by the fact that real-world outliers are clustered and not uniformly distributed in the feature space.
For example, in fault detection, outliers are typically clustered due to multiple kinds of errors. Unlike outlier sampling \cite{Barnabe-Lortie15}, exploration sampling quickly presents outliers from different clusters to an oracle by producing smaller aggregates $\agg$ for samples that lie near already sampled regions in the feature space. Consequently, an expert can faster detect different outlier causes.

Fig. \ref{fig:exploration_sampling_clusters} illustrates the incremental development of the optimal decision boundary by using exploration sampling on an artificial, two-dimensional dataset consisting of 60 targets and 60 outliers. Two OCSVMs \cite{Scholkopf2001} were used as OCCs.
Fig. \ref{fig:exploration_sampling_clusters_12_iterations} demonstrates how exploration sampling leads to uniformly exploring the feature space after 12 iterations. Finally, Fig. \ref{fig:exploration_sampling_clusters_27_iterations} shows an optimal classification result after only 27 iterations and no more than 10 presented outliers.

\subsection{Similarity sampling}
The usage of a second OCC allows to design the query strategy \emph{similarity sampling} which is similar to classical uncertainty sampling \cite{Lewis94} and differs only in the kind of uncertainty measure. Instead of using decision values from a binary classifier as uncertainty, similarity sampling uses the absolute difference between the decision values of both OCCs as uncertainty measure. The following aggregation function $\agg_{\text{sim}}$ leads to querying samples which lie in the middle between the decision regions of the target and outlier class, i.e. samples with similar values of $\DV^t$ and $\DV^o$:
\begin{equation}
	\agg_{\text{sim}} = -|\DV^t - \DV^o|.
\end{equation}
It focuses more on exploitation than on exploration.
Thereby we assume decision values that are normalized to a maximum of 1 in order to be consistent with the initial decision values in Alg. \ref{alg:proposed_al_method}.

\subsection{Stopping criterion}
In contrast to other active learning approaches, the proposed method easily allows to define the following stopping criterion:
The active learning stops if, for all samples from $\mathcal{U}$, at least one of both decision values is below its respective user-predefined threshold $T^t$ or $T^o$:
\begin{equation}
	\forall x_u\in \mathcal{U}: \DV_u^t<T^t \vee \DV_u^o<T^o.
\end{equation}
In other words, active learning is finished if for each sample at least one of both OCCs is sure enough that the sample belongs to the respective class. Regarding OCSVMs, a reasonable choice is $T^t=T^o=0$.

\begin{table}[!ht]
	\def\arraystretch{1.1}
	\setlength\tabcolsep{3.3pt}
	\caption{Datasets used in experiments}
	\label{tab:datasets}
	\centering
	\footnotesize
	\begin{tabular}{lccccc}
	\toprule
	\textbf{Dataset} & \textbf{Features} & \textbf{Targets} & \textbf{Outliers} & \textbf{Batch size} & \textbf{Iterations} \\
\midrule
	BreastW & 9 & 444 & 239 & 1 & 40 \\
	Cardio & 21 & 1655 & 176 & 1 & 40 \\
	Glass & 9 & 205 & 9 & 1 & 40 \\
	Letter & 32 & 1500 & 100 & 5 & 100 \\
	Lympho & 18 & 142 & 6 & 1 & 10 \\
	Mammography & 6 & 10923 & 260 & 30 & 50 \\
	Mnist & 100 & 6903 & 700 & 10 & 40 \\
	Musk & 166 & 2965 & 97 & 1 & 60 \\
	Optdigits & 64 & 5066 & 150 & 5 & 50 \\
	Pendigits & 16 & 6714 & 156 & 1 & 50 \\
	Satellite & 36 & 4399 & 2036 & 10 & 50 \\
	Thyroid & 6 & 3679 & 93 & 1 & 50 \\
	Vowels & 12 & 1406 & 50 & 3 & 50 \\
	WBC & 30 & 357 & 21 & 1 & 20 \\
	Wine & 13 & 119 & 10 & 1 & 20 \\
	\bottomrule
\end{tabular}
\end{table}

\begin{table*}[!ht]
	\renewcommand{\arraystretch}{1.22}
	\def\linedist{0cm}
	\setlength\tabcolsep{3.3pt}
	\caption{Experimental results (mean and standard deviation of BACC calculated over 10-fold cross-validation)}
	\label{tab:experimental_results}
	\centering
	\footnotesize
	\begin{tabular}{lcccccccc}
	\toprule
	\textbf{Dataset} & \textbf{Initial} & \textbf{Random} & \textbf{\makecell{Uncertainty \\(lh) \cite{Juszczak03}}} & \textbf{\makecell{Expected \\margin \cite{Ghasemi112}}} & \textbf{Entropy \cite{Ghasemi112}} & \textbf{Outlier \cite{Barnabe-Lortie15}} & \textbf{Similarity} & \textbf{Exploration} \\
	\midrule
	BreastW & \makecell{0.925 {\scriptsize($\pm$ 0.023)}} & \makecell{0.938 {\scriptsize($\pm$ 0.014)}} & \makecell{0.925 {\scriptsize($\pm$ 0.024)}} & \makecell{0.931 {\scriptsize($\pm$ 0.013)}} & \makecell{0.936 {\scriptsize($\pm$ 0.013)}} & \makecell{0.925 {\scriptsize($\pm$ 0.024)}} & \makecell{0.944 {\scriptsize($\pm$ 0.016)}} & \textbf{\makecell{0.953 {\scriptsize($\pm$ 0.010)}}} \\[\linedist]
	Cardio & \makecell{0.842 {\scriptsize($\pm$ 0.014)}} & \makecell{0.860 {\scriptsize($\pm$ 0.012)}} & \makecell{0.846 {\scriptsize($\pm$ 0.016)}} & \makecell{0.863 {\scriptsize($\pm$ 0.014)}} & \makecell{0.862 {\scriptsize($\pm$ 0.015)}} & \makecell{0.856 {\scriptsize($\pm$ 0.015)}} & \makecell{0.844 {\scriptsize($\pm$ 0.015)}} & \textbf{\makecell{0.878 {\scriptsize($\pm$ 0.016)}}} \\[\linedist]
	Glass & \makecell{0.600 {\scriptsize($\pm$ 0.031)}} & \makecell{0.762 {\scriptsize($\pm$ 0.027)}} & \makecell{0.681 {\scriptsize($\pm$ 0.036)}} & \makecell{0.758 {\scriptsize($\pm$ 0.023)}} & \makecell{0.755 {\scriptsize($\pm$ 0.027)}} & \makecell{0.817 {\scriptsize($\pm$ 0.018)}} & \makecell{0.792 {\scriptsize($\pm$ 0.031)}} & \textbf{\makecell{0.823 {\scriptsize($\pm$ 0.014)}}} \\[\linedist]
	Letter & \makecell{0.603 {\scriptsize($\pm$ 0.016)}} & \makecell{0.765 {\scriptsize($\pm$ 0.026)}} & \makecell{0.784 {\scriptsize($\pm$ 0.026)}} & \makecell{0.747 {\scriptsize($\pm$ 0.030)}} & \makecell{0.748 {\scriptsize($\pm$ 0.031)}} & \textbf{\makecell{0.799 {\scriptsize($\pm$ 0.014)}}} & \makecell{0.736 {\scriptsize($\pm$ 0.017)}} & \textbf{\makecell{0.799 {\scriptsize($\pm$ 0.014)}}} \\[\linedist]
	Lympho & \makecell{0.906 {\scriptsize($\pm$ 0.054)}} & \makecell{0.938 {\scriptsize($\pm$ 0.023)}} & \makecell{0.972 {\scriptsize($\pm$ 0.015)}} & \makecell{0.941 {\scriptsize($\pm$ 0.041)}} & \makecell{0.941 {\scriptsize($\pm$ 0.041)}} & \textbf{\makecell{0.985 {\scriptsize($\pm$ 0.009)}}} & \makecell{0.973 {\scriptsize($\pm$ 0.029)}} & \makecell{0.984 {\scriptsize($\pm$ 0.010)}} \\[\linedist]
	Mammography & \makecell{0.729 {\scriptsize($\pm$ 0.007)}} & \makecell{0.767 {\scriptsize($\pm$ 0.011)}} & \makecell{0.748 {\scriptsize($\pm$ 0.010)}} & \makecell{0.763 {\scriptsize($\pm$ 0.013)}} & \makecell{0.768 {\scriptsize($\pm$ 0.018)}} & \makecell{0.783 {\scriptsize($\pm$ 0.015)}} & \makecell{0.773 {\scriptsize($\pm$ 0.019)}} & \textbf{\makecell{0.786 {\scriptsize($\pm$ 0.017)}}} \\[\linedist]
	Mnist & \makecell{0.817 {\scriptsize($\pm$ 0.013)}} & \makecell{0.830 {\scriptsize($\pm$ 0.014)}} & \makecell{0.822 {\scriptsize($\pm$ 0.014)}} & \makecell{0.831 {\scriptsize($\pm$ 0.016)}} & \makecell{0.831 {\scriptsize($\pm$ 0.016)}} & \makecell{0.837 {\scriptsize($\pm$ 0.014)}} & \makecell{0.794 {\scriptsize($\pm$ 0.016)}} & \textbf{\makecell{0.851 {\scriptsize($\pm$ 0.012)}}} \\[\linedist]
	Musk & \makecell{0.779 {\scriptsize($\pm$ 0.012)}} & \makecell{0.821 {\scriptsize($\pm$ 0.017)}} & \makecell{0.779 {\scriptsize($\pm$ 0.013)}} & \makecell{0.818 {\scriptsize($\pm$ 0.017)}} & \makecell{0.817 {\scriptsize($\pm$ 0.012)}} & \makecell{0.797 {\scriptsize($\pm$ 0.015)}} & \makecell{0.797 {\scriptsize($\pm$ 0.015)}} & \textbf{\makecell{0.856 {\scriptsize($\pm$ 0.016)}}} \\[\linedist]
	Optdigits & \makecell{0.629 {\scriptsize($\pm$ 0.007)}} & \makecell{0.711 {\scriptsize($\pm$ 0.007)}} & \makecell{0.686 {\scriptsize($\pm$ 0.009)}} & \makecell{0.702 {\scriptsize($\pm$ 0.007)}} & \makecell{0.702 {\scriptsize($\pm$ 0.007)}} & \makecell{0.734 {\scriptsize($\pm$ 0.010)}} & \makecell{0.700 {\scriptsize($\pm$ 0.006)}} & \textbf{\makecell{0.755 {\scriptsize($\pm$ 0.011)}}} \\[\linedist]
	Pendigits & \makecell{0.937 {\scriptsize($\pm$ 0.006)}} & \makecell{0.941 {\scriptsize($\pm$ 0.006)}} & \makecell{0.937 {\scriptsize($\pm$ 0.006)}} & \makecell{0.943 {\scriptsize($\pm$ 0.005)}} & \makecell{0.943 {\scriptsize($\pm$ 0.005)}} & \makecell{0.940 {\scriptsize($\pm$ 0.006)}} & \makecell{0.937 {\scriptsize($\pm$ 0.006)}} & \textbf{\makecell{0.967 {\scriptsize($\pm$ 0.003)}}} \\[\linedist]
	Satellite & \makecell{0.694 {\scriptsize($\pm$ 0.015)}} & \makecell{0.723 {\scriptsize($\pm$ 0.009)}} & \makecell{0.694 {\scriptsize($\pm$ 0.016)}} & \makecell{0.703 {\scriptsize($\pm$ 0.011)}} & \makecell{0.709 {\scriptsize($\pm$ 0.013)}} & \makecell{0.697 {\scriptsize($\pm$ 0.016)}} & \makecell{0.709 {\scriptsize($\pm$ 0.017)}} & \textbf{\makecell{0.747 {\scriptsize($\pm$ 0.010)}}} \\[\linedist]
	Thyroid & \makecell{0.898 {\scriptsize($\pm$ 0.009)}} & \makecell{0.900 {\scriptsize($\pm$ 0.011)}} & \makecell{0.899 {\scriptsize($\pm$ 0.008)}} & \makecell{0.903 {\scriptsize($\pm$ 0.013)}} & \makecell{0.905 {\scriptsize($\pm$ 0.012)}} & \makecell{0.900 {\scriptsize($\pm$ 0.009)}} & \makecell{0.871 {\scriptsize($\pm$ 0.025)}} & \textbf{\makecell{0.909 {\scriptsize($\pm$ 0.011)}}} \\[\linedist]
	Vowels & \makecell{0.735 {\scriptsize($\pm$ 0.018)}} & \makecell{0.839 {\scriptsize($\pm$ 0.021)}} & \makecell{0.880 {\scriptsize($\pm$ 0.015)}} & \makecell{0.842 {\scriptsize($\pm$ 0.022)}} & \makecell{0.837 {\scriptsize($\pm$ 0.023)}} & \makecell{0.938 {\scriptsize($\pm$ 0.019)}} & \makecell{0.835 {\scriptsize($\pm$ 0.020)}} & \textbf{\makecell{0.940 {\scriptsize($\pm$ 0.015)}}} \\[\linedist]
	WBC & \makecell{0.810 {\scriptsize($\pm$ 0.026)}} & \makecell{0.839 {\scriptsize($\pm$ 0.034)}} & \makecell{0.822 {\scriptsize($\pm$ 0.033)}} & \makecell{0.852 {\scriptsize($\pm$ 0.049)}} & \makecell{0.844 {\scriptsize($\pm$ 0.043)}} & \makecell{0.874 {\scriptsize($\pm$ 0.023)}} & \makecell{0.828 {\scriptsize($\pm$ 0.060)}} & \textbf{\makecell{0.882 {\scriptsize($\pm$ 0.023)}}} \\[\linedist]
	Wine & \makecell{0.581 {\scriptsize($\pm$ 0.035)}} & \makecell{0.737 {\scriptsize($\pm$ 0.034)}} & \makecell{0.690 {\scriptsize($\pm$ 0.049)}} & \makecell{0.728 {\scriptsize($\pm$ 0.059)}} & \makecell{0.743 {\scriptsize($\pm$ 0.042)}} & \makecell{0.798 {\scriptsize($\pm$ 0.037)}} & \makecell{0.713 {\scriptsize($\pm$ 0.043)}} & \textbf{\makecell{0.815 {\scriptsize($\pm$ 0.043)}}} \\[\linedist]
\bottomrule
\end{tabular}

\end{table*}

\section{Experiments}
\subsection{Setup}
We evaluated our proposed active learning strategies on 15 datasets which are listed in Table \ref{tab:datasets}. All datasets originate from the Outlier Detection DataSets (ODDS) Library \cite{Rayana16}.
Some of them contain many possibly representative outliers to deploy binary classification. Nevertheless, we used all datasets for one-class classification. The selected datasets differ in its absolute number of samples, number of features and relative amount of outliers.

We deployed an OCSVM \cite{Scholkopf2001} with radial basis function kernel as OCC. Its optimal hyperparameters were determined by a grid search based on splitting a dataset into training and validation set. Both OCCs in the proposed active learning method were trained using the same hyperparameters.

Our evaluation procedure was similar to \cite{Barnabe-Lortie15}. We performed a 10-fold cross-validation.
The first fold was used as initial training set, the folds 2-6 were considered as pool of unlabeled samples and the remaining four folds were used as test set. The split into 10 folds was only done once.
The mapping between fold number and real fold was circularly shifted 10 times. In order to start all active learning strategies with the same amount of information, the outliers in the initial training set were ignored.

Besides the three proposed query strategies, our experiments considered the active learning approaches "lh" uncertainty sampling \cite{Juszczak03}, expected margin sampling \cite{Ghasemi112}, entropy sampling \cite{Ghasemi112}, outlier sampling \cite{Barnabe-Lortie15} and random sampling.
The two kernel density estimation strategies of \cite{Ghasemi112} were implemented using the publicly available code by Ghasemi \cite{GhasemiGithub}.
In the case of large datasets, batches consisting of multiple samples were queried after each iterative training step in order to reduce the computational effort. The batch size for each dataset is listed in Table \ref{tab:datasets}.

The metric balanced accuracy (BACC) 
\begin{equation}
\text{BACC} = \frac{1}{2}\left(\frac{\TN}{N}+\frac{\TP}{P}\right)
\end{equation}
was calculated to evaluate the results. $N$ denotes the number of targets and $P$ the number of outliers. $\TN$ represents the number of true negatives, i.e. the number of correctly classified targets. $\TP$ denotes the number of true positives, i.e. the number of correctly classified outliers.

Attention should be paid to the evaluation of active learning strategies because their performance depends on the number of queries. Due to the practical relevance, we compared query strategies after a few iterations. Thus, the number of presented samples was manually chosen in such a way that the performance improved noticeably, but did not yet saturate. The total number of presented samples equals the number of iterations times the batch size. These values are listed in \mbox{Table \ref{tab:datasets}}.

\begin{figure}[!t]
	\centering
	\setlength\figureheight{4cm}
	\setlength\figurewidth{0.4\textwidth}
	\input{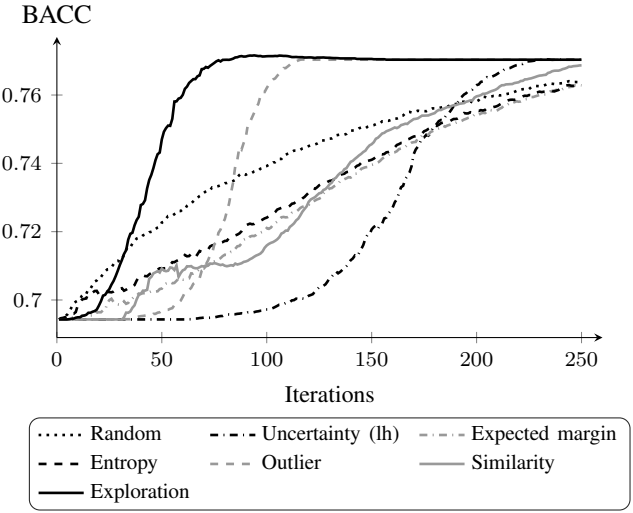}
	\caption{Balanced accuracy over number of iterations on Satellite dataset using various active learning strategies}
	\label{fig:iterations_plot}
\end{figure}

\subsection{Results}
At first, we compared all query strategies depending on the number of iterations by taking the exemplary dataset Satellite. Fig. \ref{fig:iterations_plot} shows the resulting graph.
The plot reveals that exploration sampling performs best after a few iterations until the performance saturates. Thus, exploration plays an important role at the beginning, when only few labeled samples are available.
Expected margin and entropy sampling lead both to an approximately linear increase of the BACC on the exemplary dataset. Random sampling performed best for very few iterations and was always better than expected margin and entropy sampling. Similarity sampling performed worse than random sampling for almost all considered iterations. A possible reason is that it focuses on exploitation which seems to be not as important as exploration in one-class classification. Outlier sampling performed poorly for few iterations. This is because it found mostly true outliers which have no impact on the target OCC used for classification. After enough iterations, its performance improved distinctly, because informative targets were found which contribute to a high exploration.
Finally, "lh" uncertainty sampling lead to a bad performance indicating a low exploration and exploitation. One reason for this is the fact that estimated targets with low confidence contribute little to iterative learning. Furthermore, estimated outliers with high confidences are often true outliers which do not affect the classification result.

Table \ref{tab:experimental_results} shows the results of the experiments on all datasets by listing the mean and standard deviation of the BACC calculated over all folds. The table evaluates the query strategies after a practically relevant number of iterations during which exploration is most important. The results confirm that exploration sampling performed best for most of the datasets.
Furthermore, it appears that outlier sampling achieves a slow exploration for datasets with relatively many outliers like BreastW or Satellite due to our previous considerations. If only few outliers are present, its performance can be comparable to or even slightly better than exploration sampling. As expected, similarity sampling generally performed bad, because it leads to a low exploration.
The remaining query strategies could not compete with exploration and outlier sampling either.

\section{Conclusion}
We presented a new active learning method for one-class classification. It is able to take labeled outliers into account by employing a second OCC. Hence, our method can develop its full potential in applications in which some outliers are available and one-class classification is preferred over binary classification. The decision values of both OCCs are combined through an aggregation function which is chosen based on a desired trade-off between exploration and exploitation. We proposed two example aggregation functions which focus more on exploration and exploitation, respectively.
The experiments showed that exploration sampling outperforms the other active learning strategies for almost all datasets after a practically relevant number of queries.
Thus, exploration seems to be more important than exploitation in active learning for one-class classification.
Future implications of this paper may include research on combining or applying other aggregation functions and using different hyperparameters for both OCCs. Finally, it is conceivable to apply the method to binary or multi-class classification.

\bibliographystyle{IEEEbib}
\bibliography{refs}
\end{document}